\pdfoutput=1

\documentclass[11pt]{article}

\usepackage[final]{acl}
\usepackage{tabularx}
\usepackage{times}
\usepackage{latexsym}
\usepackage[subtle]{savetrees}

\usepackage[T1]{fontenc}

\usepackage[utf8]{inputenc}

\usepackage{microtype}

\usepackage{inconsolata}

\usepackage{graphicx}
\usepackage{booktabs}
\usepackage{siunitx}
\usepackage{bbm}
\usepackage{bm}
\usepackage[inline]{enumitem}
\usepackage{amsmath,amssymb}
\usepackage{multirow}
\usepackage[export]{adjustbox}
\usepackage{array}
\usepackage{longtable}

\newcommand{\nl}[1]{\textit{``#1''}}

\newcommand{\lv}{\texttt{LLaVA}}
\newcommand{\ls}{\texttt{LLaVA7B}}
\newcommand{\lb}{\texttt{LLaVA34B}}
\newcommand{\lm}{\texttt{LLaVA-MORE-8B}}
\newcommand{\clip}{\texttt{CLIP}}
\newcommand{\sig}{\texttt{SigLIP}}
\newcommand{\lmc}{\texttt{LM-\clip{}}}
\newcommand{\lms}{\texttt{LM-\sig{}}}
\newcommand{\qwen}{\texttt{Qwen2-VL}}
\newcommand{\pop}{%
  \leavevmode
  \mbox{
    {\normalsize P}
    {\small OP}
    {\normalsize VQA}
  }%
}

\title{Performance Gap in Entity Knowledge Extraction\\Across Modalities in Vision Language Models}

\usepackage{authblk}
\author{Ido Cohen ~~~~ Daniela Gottesman ~~~~ Mor Geva ~~~~ Raja Giryes\vspace{7pt} \\
  Tel Aviv University \vspace{7pt} \\
  \small \texttt{ \{idoc@mail, gottesman3@mail, morgeva@tauex, raja@tauex\}.tau.ac.il}
}

\begin{document}
\maketitle
\begin{abstract}
Vision-language models (VLMs) excel at extracting and reasoning about information from images. Yet, their capacity to leverage internal knowledge about specific entities remains underexplored. This work investigates the disparity in model performance when answering factual questions about an entity described in text versus depicted in an image. Our results reveal a significant accuracy drop — reaching 18\% for some models — when the entity is presented visually instead of textually. To study this gap we present \pop{}, a dataset which allows separating entity recognition and question answering, and use it to benchmark several models. We hypothesize that this decline arises from limitations in how information flows from image tokens to query tokens. Thus, we use mechanistic interpretability tools to reveal that, although image tokens are preprocessed by the vision encoder, meaningful information flow from these tokens occurs only in the much deeper layers. Furthermore, critical image processing happens in the language model's middle layers, allowing few layers for consecutive reasoning, highlighting a potential inefficiency in how the model utilizes its layers for reasoning. These insights shed light on the internal mechanics of VLMs and offer pathways for enhancing their reasoning capabilities. \pop{} can be found \href{https://huggingface.co/datasets/idoco/PopVQA}{at this link}.

\end{abstract}

\begin{figure*}[t]
    \centering
    \includegraphics[width=0.9\textwidth]{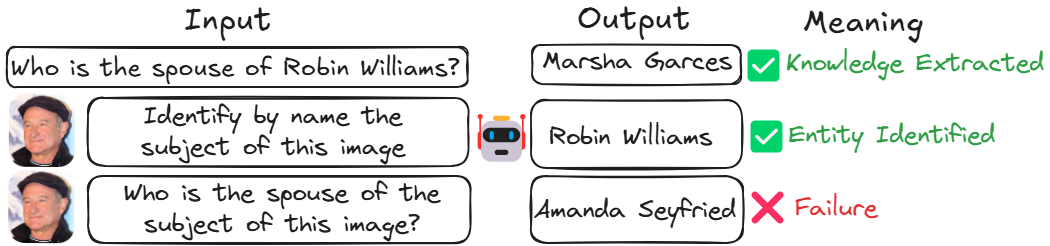}
    \caption{Illustration of our motivating observation. The model successfully extracts its knowledge about the entity when the entity is represented in text. It further successfully identifies the entity in the image, but it fails to combine the two.}
    \label{fig:teaser}
\end{figure*}

\section{Introduction}
Over the past several years, vision-language models (VLMs) have made remarkable progress in linking visual inputs with natural language descriptions, enabling them to perform tasks such as image recognition \cite{clip, mpvr}, video understanding~\cite{lin2023video}, and Visual Question Answering (VQA)~\cite{liu2023visual, llava-next}. Much of the existing research evaluating these models focuses on their capacity to understand and describe information presented in the input image: counting objects, assessing complex visual arrangements, and understanding interactions between humans and objects \cite{dang2024towards,Basu2024}. While this has demonstrated that VLMs can integrate visual and textual information effectively, it does not fully address a critical question: 

\textit{how well can VLMs exploit their internal knowledge about entities presented in an image rather than ones simply described in text?}

This question opens a new angle on VLM evaluation. Instead of probing their grasp of broad scenes, we consider a different scenario: retrieving factual knowledge about visually presented entities. For example, if the subject of a question is a notable figure, say Robin Williams, we can ask the model \textit{"Who is Robin William's spouse?"}, which we refer to as a textual representation. Or we can provide the model with an image of Robin Williams and ask \textit{“Who is the spouse of the subject in this image?”}, which we refer to as a visual representation. The second case may be viewed as a two-hop reasoning task. First, the model must recognize the entity visually—\textit{identification}. Next, it must connect this recognized entity to its stored factual knowledge—\textit{extraction}. Figure \ref{fig:teaser} provides an example where the model contains the knowledge to answer the question, but fails to extract it when provided with a visual representation. A detailed explanation of this evaluation appears in Section \ref{sec:gap}.

Our empirical findings reveal a large accuracy drop when relying on visual representations rather than textual ones. Further probing suggests that this stems from the model investing too many layers in identifying the subject visually, leaving fewer layers for the second hop of factual reasoning \citep{biran-etal-2024-hopping}.
To investigate these dynamics more concretely, we utilize techniques of mechanistic interpretability such as activation patching \citep{NEURIPS2020_92650b2e} and attention knockout \citep{geva-etal-2023-dissecting}. They allow us to intervene in the model’s processing mid-inference, altering how and when image information influences the final answer. By tracking the flow of information through the network, we show that entity recognition and knowledge extraction are deeply entangled processes, with identification dominating the representational space longer than what one would expect. Consequently, the model’s internal knowledge is leveraged less efficiently.

The contribution of our work is twofold. First, we release \pop, a dataset that facilitates validating entity identification prior to question answering. Second, we provide a better understanding of how VLMs process entities and integrate their internal factual knowledge with image information, highlighting a gap between the processing of visual and textual information. Instead of treating entity-level knowledge extraction as a monolithic process, we reveal that it involves at least two interdependent steps. We show that the initial reliance on visual representations can restrict later factual reasoning compared to textual representations. This understanding highlights the need for more multi-modality specific evaluation protocols. 

Our experiments, which measure the performance gap and apply activation-based hidden state interventions, provide a foundation for future efforts to enhance the models’ ability to draw on their underlying knowledge when presented with visual inputs. Code to reproduce our experiments can be found \href{https://github.com/ido-co/vlm-modality-gap}{at this link}.


\section{Problem Setup}

\paragraph{Goal.} We ask how well VLMs can leverage their internal factual knowledge when presented with visual representations of entities, compared to textual ones. We suggest that VLMs perform two key steps: identifying the entity in the image, followed by  connecting it to its internal knowledge to provide factual information. By isolating these steps and examining where models might fall short, we aim to pinpoint the layer-by-layer flow of information and advance our understanding about why models struggle to draw on their internal knowledge when relying on visual inputs.

\paragraph{Preliminaries and Notation.} 
VLMs are architectures that integrate visual and textual modalities to perform tasks such as image captioning, visual question answering and more. Our VLM analysis in this work is mainly focused on the common architecture of \lv{}  \cite{liu2023visual}, and thus, we will follow the same notation.

\begin{figure*}[t]
    \centering
    \includegraphics[width=0.98\textwidth]{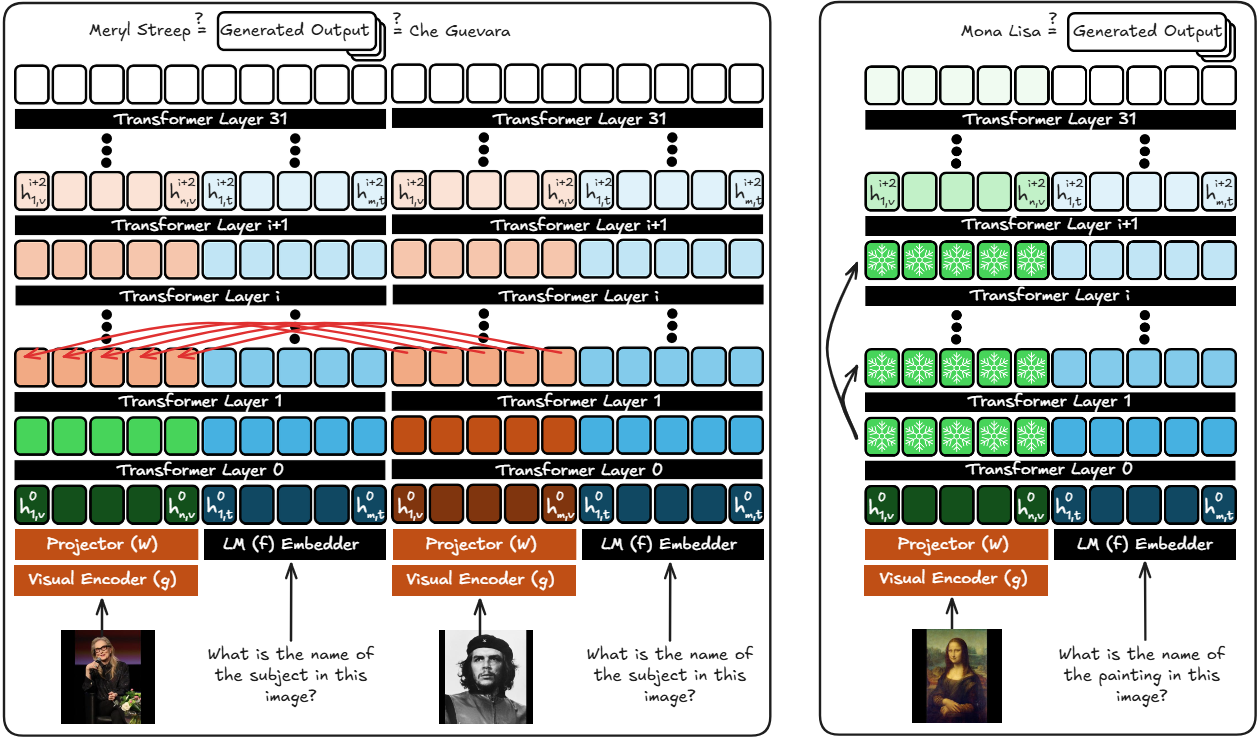}
    \caption{The left pane is an illustration for the experiment in Section \ref{sec:cross}. The hidden states at the positions of the visual tokens are copied from the forward pass of the injected entity at the source layer (in this case layer 1) replacing the hidden states of the original entity's forward pass at the same layer and position. The right pane is an illustration of the experiment in Section \ref{sec:forward}. Starting from the source layer, the hidden states at the positions of the visual tokens are copied and patched instead of the hidden states at the same positions in the subsequent layers until the target layer, effectively "freezing" their processing by the LLM.}
    \label{fig:exp}
\end{figure*}

The VLM is composed of a pre-trained visual encoder $g$, a projection layer $W$, and an LLM $f$. Given an input image $\mathbf{X}_\mathbf{v}$ and text query $\mathbf{X}_\mathbf{t}$, $\mathbf{X}_\mathbf{v}$ is first processed by $g$ to provide the visual features $\mathbf{Z}_\mathbf{v}=g(\mathbf{X}_\mathbf{v})$ then passed through the projection layer to align the visual features into the LLM's embedding space as a series of tokens $\mathbf{H}_\mathbf{v}=W\cdot\mathbf{Z}_\mathbf{v}$. In parallel, $\mathbf{X}_\mathbf{t}$ is passed through the LLM's tokenizer and embedding layer to create a series of tokens $\mathbf{H}_\mathbf{t}$. Next the two are concatenated and passed through the LLM $f([\mathbf{H}_\mathbf{v}, \mathbf{H}_\mathbf{t}])$.
Figure~\ref{fig:exp}, which illustrates the experiments we perform in the paper (see more details in Section~\ref{sec:cross} and \ref{sec:forward}), can be used also to better understand these notations.

Every token in this input can be written as $\mathbf{h}_{i,v}^{0}$ or $\mathbf{h}_{i, t}^{0}$ where the $v$ and $t$ indicate a visual or textual token, the $i$ indicates the $i^\text{th}$ visual or textual tokens, and the $0$ indicates input to layer $0$. As the language model processes the entire input at each layer $l$, it builds a representation $\left(\mathbf{h}_{1, v}^{\ell} \ldots \mathbf{h}_{n, v}^{\ell}, \mathbf{h}_{1, t}^{\ell} \dots \mathbf{h}_{m, t}^{\ell}, \mathbf{h}_{1, g}^{\ell} \dots \mathbf{h}_{k, g}^{\ell} \right)$, where $\mathbf{h}_{1, g}^{\ell}...\mathbf{h}_{k, g}^{\ell}$ are the generated tokens hidden representations, $\left| \mathbf{H}_\mathbf{v} \right| = n$, $\left| \mathbf{H}_\mathbf{t} \right|=m$ and $\ell$ is the layer. We refer to the tokens once inside the LLM as hidden states.

In our experiments we focus on how information flows from the visual hidden states $\left(\mathbf{h}_{1, v}^{\ell} \ldots \mathbf{h}_{n, v}^{\ell}\right)$ to the textual and generated tokens. We do this by applying techniques of mechanistic interpretability.

\paragraph{Dataset and Experiments Overview.}
To analyze how VLMs integrate visual and textual information, we introduce a new dataset called \pop{} in Section~\ref{sec:popVQA}, featuring popular entities that VLMs are likely to have internal knowledge about. This dataset focuses on a two-hop reasoning process: (i) identification of an entity based on an image, followed by (ii) extraction of factual information related to that entity. 
We compare model responses when the entity is given as text (“Robin Williams”) versus as an image, providing insight into the gap between visual and textual representations. In these experiments, we first present a motivational case study in Section~\ref{sec:gap}, demonstrating that various VLMs (e.g., \ls, \qwen) show a marked accuracy drop when relying on visual inputs. This experiment identifies a gap between the use of visual and language information in VLMs.

Motivated by the first experiment, we turn to further explore how VLMs rely on visual and textual information. We delve deeper into the reasons behind the gap, focusing on \lv{} with different encoders such as \clip{} and \sig{}.
Specifically, we aim to understand at what point in the inference pass the entity in an image is identified, and this identity is propagated to the query and generated token positions.

Through activation patching \citep{NEURIPS2020_92650b2e} and other mechanistic interpretability methods, we trace how and where entity recognition occurs, and how it impacts the model’s ability to retrieve the correct facts.
Section~\ref{sec:cross} describes how we employ activation patching to identify the layer at which the model extracts the entity identity, establishing an upper bound on identification layers. Building on these findings, Section~\ref{sec:forward} applies forward activation patching to investigate whether the entity identity could be inferred at earlier layers. 

We include additional experiments using attention knockouts \citep{geva-etal-2023-dissecting} in appendix \ref{sec:knockout}, which further validate our conclusions. Throughout the paper, we use \pop{} to systematically evaluate VLM performance on visually driven factual retrieval and demonstrate how visual and textual components interact within a unified transformer architecture.


\section{The \pop{} Dataset}
\label{sec:popVQA}

\begin{figure}
    \centering
    \includegraphics[width=0.98\linewidth]{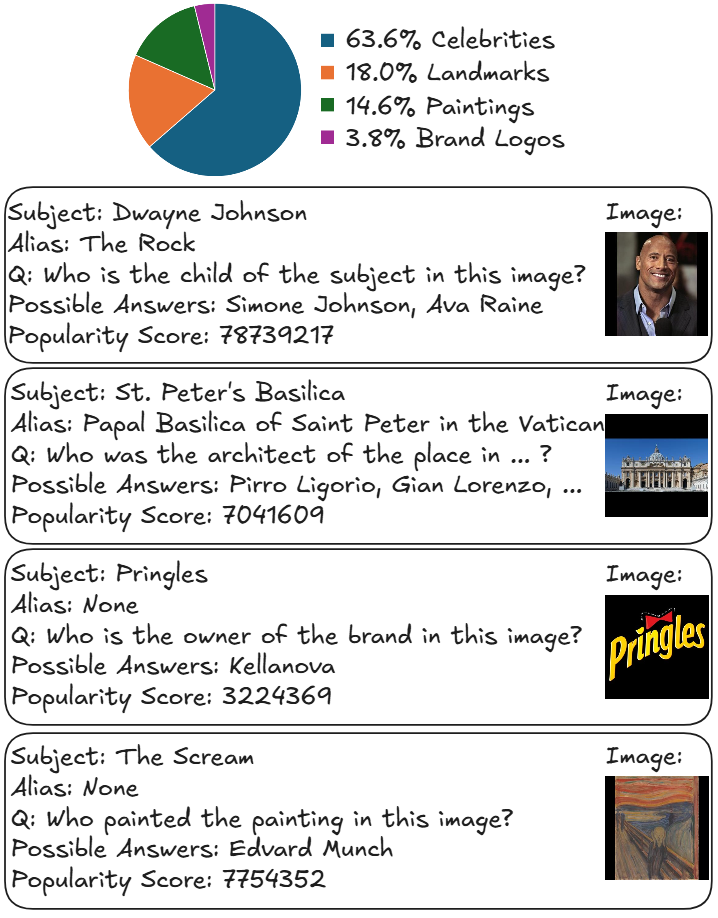}
    \caption{The \pop{} dataset contains various entity-image pairs of celebrities, landmarks, brand logos and painting. For each entity we keep the image, its subject, popularity score, possible aliases and questions and answers related to that entity.}
    \label{fig:pop}
\end{figure}

In order to isolate the model's ability to extract knowledge about a given entity from its ability to identify it in the first place, we want to restrict our questions to entities that the model recognizes. Other datasets that have knowledge seeking goals such as infoSeek \citep{chen-etal-2023-pre-trained} or VeQuAE \citep{lerner2022viquae} do not include the entity in the image but rather only the answer to the question, thus, making it difficult to do this separation. Therefore, we built the \pop{} dataset to facilitate this need. It is inspired by PopQA \cite{mallen-etal-2023-trust} that created a QA dataset for various entities using Wikipedia.

The first part of \pop{} comprises of 15,395 entity-image pairs from Wikidata \citep{vrandevcic2014wikidata} along with aliases and popularity score (number of page views).
The vast majority of entities are celebrities (63.6\%). To allow ablation of other entity types, we added landmarks (18\%), such as famous structures or UNESCO sites, famous paintings (14.6\%), and brand logos (3.8\%). All images are resized to 336 × 336 which is a standard input resolution, by adding black borders to create a square, then resizing. This was done in order to maintain the original aspect ratio, without losing any details. The popularity score was used only in the creation of the dataset such that we select only entities with high popularity, which increases the chance that the models will recognize them \cite{mallen-etal-2023-trust}. We do not use this value directly in the experiments. 

The second part of \pop{} provides information seeking questions about the entities; at least two per entity. To generate the questions, we convert subject-relation-object triplets from Wikidata into question-answer pairs using manually designed templates \citep[e.g.,][]{petroni-etal-2019-language, sciavolino-etal-2021-simple}. For example, the triplet $(\textit{Robin Williams, spouse, Marsha Garces})$ is transformed into the question, \nl{Who is the spouse of Robin Williams?}, with the answer \nl{Marsha Garces}. We filter out questions where the answer can be deduced from the image alone without recognizing the entity, such as eye color. We point out that this is not bulletproof as, for example, the native language of an athlete can be guessed if the athlete is wearing a national team jersey.

One legitimate concern is the presence of stale information, particularly for entities like celebrities whose personal circumstances (e.g., marital status, offspring) may change over time. However, this issue affects only a small subset of the entities in \pop{}, and an even smaller fraction of the associated questions. To mitigate this, we allow multiple correct answers per question where applicable. Additionally, we provide code for rebuilding \pop{} using up-to-date snapshots of Wikipedia and Wikidata, enabling users to generate a version that reflects more recent factual knowledge.

In the textual representation case, the model is presented with the question as is. For the visual case, the model receives an image of the entity along with a reformulated query, such as \nl{Who is the spouse of the subject of this image?}. This ensures that the model needs to internally identify the subject depicted in the image before reasoning and generating an answer.

\section{Motivating Experiment}
\label{sec:gap}
We start by quantifying the model's performance gap in entity knowledge extraction when presented with visual versus textual representations of entities. Given that the model recognizes the entity, asking about the visual representation, is not a harder task in essence, but a matter of computational capacity. Our motivating experiment demonstrate that, indeed, using the visual representation is more challenging. The subsequent experiments aim at explaining this gap and suggest that the reason for this bigger challenge is that the model needs to use part of its capacity to first identify the subject in the image and only then it can answer the question with the remaining capacity.
Thus, as the model is of a fixed depth, once the model identifies the entity, it becomes a question of extracting the answer within the limits of the `remaining compute' of the model. 

Given the above discussion, we relate to the textual representation accuracy as the potential of the model. The lower the drop between the textual and visual accuracy, the better the model is at utilizing the visual input to elicit extraction of its internal knowledge. This could indicate better alignment between the embedding spaces of the LLM and the visual backbone.

To run the experiment, we first identify which entities each model is capable of recognizing. This is done by presenting the model with an image of each entity along with a prompt asking it to identify the subject (e.g., “Identify by name the subject in this image.”). Entities for which the model returns the correct name are considered identified. We then evaluate the model's ability to answer factual questions for these identified entities given the two representations, textual versus visual, using the formulation in Section \ref{sec:popVQA}. This setup ensures that performance differences are not due to failures in entity recognition, but reflect differences in how well the model retrieves knowledge once the entity is known.

We show results for the following models; \texttt{\lv-1.5-7B} \citep{liu2023visual} as the canonic base VLM; we refer to it as \ls{}. \lm{}  \citep{cocchi2025llavamorecomparativestudyllms} which uses a larger Llama 3.1 as its LLM and we try both variants using \clip{} \citep{Radford2021LearningTV} and \sig{}  \citep{10377550} as backbones allowing for controlled comparison between visual backbones; we refer to these as  \lmc{} and \lms{}. We use \texttt{\lv-1.6-34B} to assess the effect of model scale, and refer to it as \lb{}. We also compare to \qwen{} \citep{Qwen2-VL} that has a different architecture and training scheme.

\paragraph{Results.} 
Table \ref{tab:popvqa} shows the results for the different models.
We measured accuracy per entity as the proportion of correctly answered questions for that entity and calculated the overall mean accuracy across all entities.
To verify the significance of the results, we performed a Wilcoxon signed-rank test between the per-entity accuracies for visual versus textual representations. The p-values range from $p = 4.02 \times 10^{-29}$ for the lower drop to $p = 3.83 \times 10^{-76}$ for the higher drop.
A more fine grained analysis by the entity types can be found in appendix \ref{sec:breakdown}. 

\begin{table}[t!]
\footnotesize
\centering
\begin{tabularx}{\columnwidth}{l|lcc|c}
\toprule
Model  &   \#Identified & Img Acc & Txt Acc & \textbf{Drop} \\
\midrule
\ls{}    &      925  &     0.276 &     0.453 &  \textbf{0.177} \\
\lmc{}   &      844  &     0.216 &     0.378 &  \textbf{0.163} \\
\lms{}   &      660  &     0.201 &     0.377 &  \textbf{0.176} \\
\lb{}    &      1286 &     0.534 &     0.656 &  \textbf{0.121} \\
\qwen{}  &      3143 &     0.433 &     0.476 &  \textbf{0.043} \\
\bottomrule
\end{tabularx}
\caption{Results on \pop. Reported accuracy is first calculated per subject then averaged across the entire dataset.}
\vspace{-0.1in}
\label{tab:popvqa}
\end{table}

 Surprisingly, \ls{} outperforms both \lm{} variants, recognizing more entities and reaching higher accuracy. Both \lm{} variants have similar accuracy, both with visual and textual representations, but the variant that uses \sig{} as a visual backbone identifies considerably more entities. \lb{} has significantly better results with a lower drop showing the benefit of increasing model depth to improve the computational capacity. \qwen{} also exhibits strong results, with the lowest drop, pointing to the advantage of more modern training schemes which include unfreezing the visual backbone parameters during training.

Interestingly, a common mistake made by the model when asked to name the subject's spouse/parent/sibling/child, is to name the subject itself. This observation provides an indication to our hypothesis that by the time information about the entity propagates from the image tokens to later token positions, there is not enough computational capacity left to use this information, and therefore the model outputs the entity's name.

Note, though, that some entities did exhibit an increase in accuracy with visual inputs. This occurred specifically when textual input alone was insufficient for reasoning, but visual input provided contextually inferable details. For example, a subject wearing the French national team uniform, implicitly informed the model that the language spoken by the subject is French.

\section{Late Information Flow}
\label{sec:cross}


To analyze the information flow in the VLMs, we apply interventions in their inference. We continue to focus on entity-image pairs for which, without intervention, the model correctly identifies the entity in the image. For all the following experiments we focus on the \lv{} architecture due to its simple architecture that allows for controlled probing, leaving other architectures as future work. We use \ls{} which successfully recognizes 925 entities from \pop{}. Yet, in order to show comparative results with different backbones we show also results on \lms{} and \lmc{} as well. These backbones were chosen because their training process inherently brings visual and textual representations closer which makes them a common choice in VLMs.

We begin by identifying the point of critical information flow from the image hidden representation to later positions. This will provide an upper bound on the layer depth at which entity identification happens and point us to where this information is likely to propagate onward. We can subsequently infer that by this point the model mostly finalizes its image-based representation enrichment and focuses on leveraging this information for prediction. We hypothesize that this happens in deep layers, leaving few layers for factual knowledge extraction, which may begin to explain the performance gap presented in Section \ref{sec:gap}. 

The experiment involves two distinct forward passes through the model, using different entities: an \textbf{original entity} and a \textbf{injected entity} (as shown in Figure 
\ref{fig:exp}). Each forward pass is initiated with a prompt asking the model to identify the subject of an image. We extract the image-token representations $\mathbf{h}_{1, v}^{\ell} \ldots \mathbf{h}_{n, v}^{\ell}$ from the \textbf{injected entity} forward pass at a specific layer. These representations are then patched at the same token positions and same layer of the forward pass of the \textbf{original entity}. Hence, we call this technique cross patching.
After patching, we allow the model to complete its forward pass and predict which entity is identified. The layer at which the predicted entity switches from being the injected entity to the original entity identifies the critical point where the model has completed sufficient processing of the visual representation such that patching does not affect its prediction. When the patching happens between entities of the same type, we use the original prompt. When we patch between different type entities, we replace the reference word (painting, place, etc.) with the word "subject" so as not to bias the output to a certain type. 

\begin{figure}[t]
\centering
\includegraphics[width=1\linewidth]{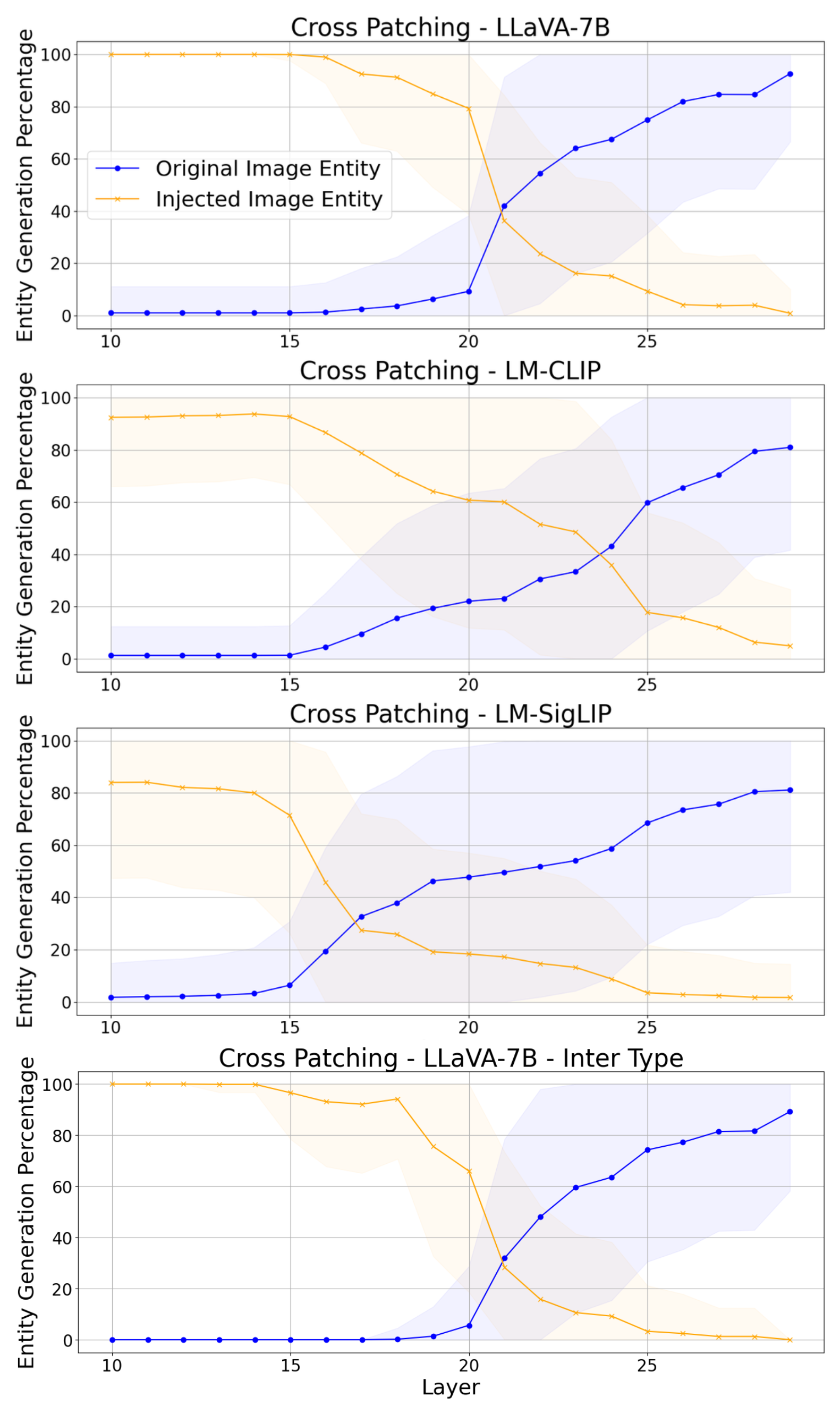}
\vspace{-0.2in}
\caption{Results for the experiment in Section \ref{sec:cross}, illustrated in Figure \ref{fig:exp}. The top three graphs show results when cross patching between forward passes of entities of the same type, for three different models (indicated in the titles). The bottom graph shows results for cross patching between different types, for \ls{}. The X axis indicates the injection layer. The orange line indicates the percent of times the injected entity was generated. The blue line is the percent of times the original entity was generated.}
\vspace{-0.1in}
\label{fig:patching}
\end{figure}

\paragraph{Results.}
Figure~\ref{fig:patching} shows the effect of patching image hidden states across different layers, indicated by the X axis. The first three graphs show results for patching between entities of the same type, for different models. The three models exhibit different behaviors but with similar trends. The main difference between the models is the layer of the critical point, mentioned earlier (where the blue and orange lines cross). Although for all models this occurs after the middle of the model (layer 16), it varies from 17 for \lms{} to 24 for \lmc{}. It is important to note the higher standard deviation for the \lm{} variants. For \ls{} the critical point is clearly marked by a sudden change at layer 20, while for \lmc{} the change from predicting the injected entity to the original is more gradual but still clearly after the middle layer.

 In all three cases, when the image tokens are patched at or before the middle layer, the model predominantly predicts the injected entity. This strengthens the hypothesis that the middle layers play a critical role in how the model interprets the image. Our results suggest that despite having 15-20 layers to process and propagate information about the original entity to later positions, the information conveyed within the narrow window of the central layers ultimately determines the outcome. Patching beyond layer 25 results in the model reliably predicting the target entity, suggesting that image-based entity identification is mostly finalized by this stage.

The last (bottom) graph in Figure~\ref{fig:patching}, shows results for patching between different types of entities for \ls{}. Note the almost identical behavior to cross patching between same types (top), albeit a slightly faster decline of the orange line. This provides even stronger evidence that the model's prediction is compelled by the information propagated in latter layers. Even when having 13 layers to attend to the visual hidden states that encode the identity of, say, LeBron James, new information that comes in at layer 14 encoding the logo of Spotify causes the model to predict the latter. Up until layer 14 this happens with a striking ~100\% of entities, regardless of type.

The results suggest two possible processing paths:

\noindent I. \textbf{Parallel identification and knowledge extraction:} Generation and query hidden states, can be enriched by the image hidden states, via the attention mechanism, concurrently with entity identification before the middle layers. This would imply that the model incrementally combines these processes over several layers,  with information passed in later layers having more impact.

\noindent II. \textbf{Sequential processing:} Identification may occur just after the middle layer, with the remaining layers allocated to factual reasoning and prediction tasks. This would leave a limited subset of late layers for extracting factual knowledge about the entity, potentially leading to degraded question-answering performance.

\section{Long Processing is Often Necessary}
\label{sec:forward}
Having identified the middle layers as the point where critical information propagates from image hidden states to later positions, and having established an upper bound on where identification happens, 
we will try to answer which of the above mentioned processing paths is more likely, and also answer whether all these layers are truly required for identification. Could it be that the entity was identified in earlier layers, but the model was unable to utilize this until later layers?

To answer these questions we apply activation patching of the image hidden states from an early layer into all subsequent layers until layer 20. In detail, at every run we choose a source layer \textit{source} and set $\left(\mathbf{h}_{1, v}^{\ell} \ldots \mathbf{h}_{n, v}^{\ell}\right) = \left(\mathbf{h}_{1, v}^{\textit{source}} \ldots \mathbf{h}_{n, v}^{\textit{source}}\right)$ for values of $\ell$ ranging from \textit{source} to 20. We repeat this for \textit{source} layers 0 to 19. This effectively "freezes" the hidden states at the image token positions, not allowing them to be updated by neither the attention nor MLP modules. Importantly, this does not prevent the query and generated positions to attend to the image tokens, only that they see the same state at all the patched layers. We choose 20 as the end point since this is past the point we identified in the last experiment, where we claim that entity identification is mostly done and propagated to later positions. If identification indeed happens in early layers, "freezing" all the layers until the point of information propagation should retain most of the identification accuracy.

\begin{figure}[t]
    \centering
    \includegraphics[width=1\linewidth]{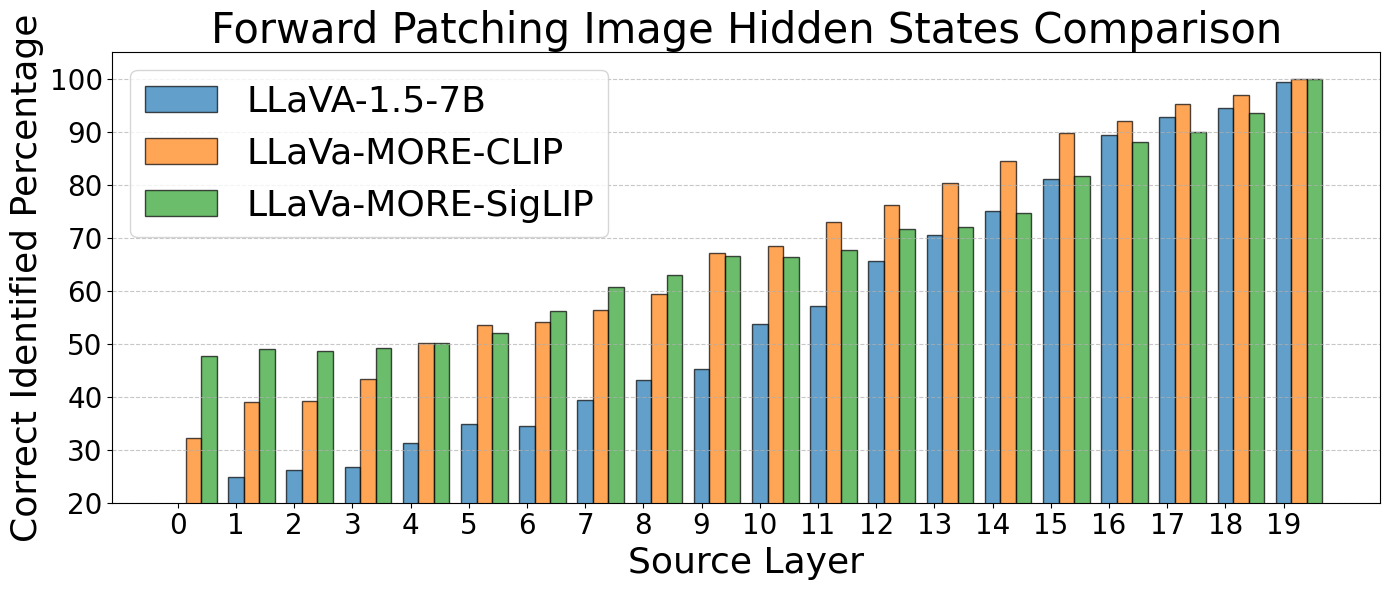}
    \vspace{-0.3in}
    \caption{Results for the experiment in Section \ref{sec:forward}, illustrated in Figure \ref{fig:exp}. Each bar shows the percent of correctly identified entities after applying patching from the layer indicated by the x axis into all layers up to layer 19.}
    \label{fig:fwd}
    \vspace{-0.1in}
\end{figure}

\paragraph{Results.} Figure \ref{fig:fwd} shows the results, where every bar shows the percent of correctly identified entities after applying patching from the \textit{source} layer indicated by the x axis onto all layers until layer 20. There is no distinction here between different types of entities. From these results we see an interesting difference between the models. For \ls{}, processing the image hidden states for more layers is required to reach the same results as \lms{} and \lmc{} do with between 4 and 10 layers less. All layers seem to take some part in this process, as every layer allows identifying more entities. As expected, this gap diminishes the more layers are allowed for processing. When patching through layers 13 to 20, \ls{} catches up to the other models. 
Interestingly, for both \lm{} variants, patching through all layers from 0 until 20, still retains almost 50\% (\lms{}) and over 30\% (\lmc{}) of recognized entities. This suggests that in these cases, the representation provided by the visual backbone encodes a sufficient signal to identify the entity. This alone would not be enough without the LLM being able to utilize this representation, suggesting better alignment of the visual and textual embedding spaces. These results complement those of the previous experiment showing that \lms{}, which had the earliest critical point in the previous experiments, can identify the most entities with the least amount of processing (manifested in number of layers).

With these results in mind, we turn back to our motivating experiment and ask whether early identification leads to better QA accuracy. As the graph shows, this question is mostly relevant for the \lm{} variants.
In this experiment, we split our dataset into entities where forward patching from early $(<5)$ layers maintains correct identification. We refer to these as \textit{early-id}, and those requiring late layers, \textit{late-id}.

Table \ref{tab:early} shows the results for both models on both splits. Notice that the early-id entities achieve higher accuracy in both models and this rise is consistent for both the image and textual representation.
This consistency between modalities teaches us that better knowledge of an entity (higher accuracy) leads to early recognition, and not the other way around. Had that been the case, we would expect to see a rise in image accuracy for early-id entities without a corresponding rise in text accuracy. This leads to the accuracy drop between image accuracy and text accuracy to actually be higher for the early-id entities. This means that although the model has access to the entity's identity at an early layer, leaving more layers for the second hop, the model does not utilize this advantage. This could indicate that although this information exists, it does not reach the query tokens via the attention mechanism early enough to inform subsequent factual recall. This strengthens the observation in Section \ref{sec:cross} that information propagates in the middle layers.

\begin{table}[t!]
\footnotesize
\centering
\begin{tabularx}{\columnwidth}{ll|cc|c}
\toprule
Model & Split & Img Acc & Txt Acc & \textbf{Drop} \\
\midrule
\lmc    & Early  & 0.260 &       0.442 &        \textbf{0.182} \\
\lmc    & Late   & 0.192 &       0.347 &        \textbf{0.154} \\
\lms    & Early  & 0.234 &       0.417 &        \textbf{0.183} \\
\lms    & Late   & 0.168 &       0.338 &        \textbf{0.170} \\
\bottomrule
\end{tabularx}

\vspace{-0.1in}
\caption{Results on \pop{}, split by early and late identified entities. Early identification is correlated with higher accuracy but not with lower drop.}
\vspace{-0.1in}
\label{tab:early}
\end{table}

\section{Related Work}
\label{sec:related}

While VLMs demonstrate impressive capabilities in tasks such as visual reasoning, object recognition, and multimodal understanding \cite{Basu2024}, their internal mechanisms and interpretability remain active areas of research. Below, we review the related works categorized into general VLM interpretability, cross-modal information flow, attention mechanisms, and mechanistic approaches to visual token processing.

\citet{dang2024towards} survey VLMs proposing a framework categorizing interpretability across data, model, and training perspectives. They highlight the alignment and decomposition of modalities as key challenges, emphasizing the need for explainability to ensure trustworthiness in multimodal AI applications. In this context, techniques for token-level, neuron-level, and embedding-level interpretability have been systematically analyzed, offering insights into the mechanisms driving decision-making processes.

Previous research has examined the internal information distribution within these models~\cite{Basu2024}, highlighted the limitations arising from pre-trained vision encoders~\cite{Tong2024}, and investigated the issue of hallucinations in VLMs~\cite{Bai2024,jiang2025interpreting}. 
Studies such as \citet{huo2024mmneuron} delve into neuron-level interpretations, uncovering how specific neurons contribute to multimodal reasoning. Similarly, \citet{jiang2024interpreting} address hallucination issues in VLMs by examining the influence of vision-language representations. Both works highlight the importance of dissecting model components to improve transparency and robustness.

On the methodological front, attention-based mechanisms remain a cornerstone for interpretability. \citet{tao2024probing} explore global and local semantic representations, while \citet{liu2023visual} focus on visual instruction tuning to enhance alignment in cross-modal embeddings.  \citet{Chefer_2021_ICCV} provides a technique to highlight the input pixels in an image that affects the decision of a transformer. \citet{Chen_2022_ACCV} further extends this approach for \clip{} and \citet{yellinek20233vlusingtreesteach} suggests using a tree structure to improve the use of HilaCAM \cite{Chefer_2021_ICCV} in VLMs.

Finally, causal approaches and model manipulation techniques have gained traction for evaluating VLMs. \citet{palit2023towards} introduce causal tracing tools to understand information flow in vision-language systems, while \citet{zhang2024from} analyze information redundancy and relevance across reasoning tasks. These works demonstrate the efficacy of manipulating hidden states and attention mechanisms to investigate and improve model performance.

Some datasets probe how well VLMs handle knowledge-intensive queries \cite{Lerner2024Cross-Modal,zhu2024unravelingcrossmodalityknowledgeconflicts,Onoe2025DOCCI,yanuka-etal-2025-bridging,Wei2025UniIR}. For instance, InfoSeek \cite{chen-etal-2023-pre-trained} is a VQA dataset tailored for information-seeking questions that cannot be resolved using only common sense knowledge. Their findings reveal that state-of-the-art VLMs often struggle with these questions; however, fine-tuning on InfoSeek enables them to leverage more fine-grained knowledge learned during pre-training. Unlike InfoSeek, which focuses on direct information-seeking queries, our dataset explicitly includes a bridge entity as part of a two-hop reasoning process—requiring both identification of the entity and extraction of the relevant fact. This design allows us to disentangle the efficacy of visual recognition from the subsequent factual retrieval and more directly analyze the gap between textual and visual representations.

Our work is most related to \citet{neo2025towards,kaduri2024whats} who investigated processing of information in VLMs. 
\citet{neo2025towards} provided insights into the role of attention layers in capturing visual attributes and object-specific details.
Contemporary to us, \citet{kaduri2024whats} demonstrated that query tokens store high-level image descriptors. Their findings reveal that middle layers of the model are critical for cross-modal knowledge transfer, while fine-grained visual details are extracted in a spatially localized manner. 
Yet, both of these works do not study the gap we reveal between the processing of textual and visual representations.

\citet{parcalabescu-frank-2023-mm} introduce a method that uses Shapley values to quantify how VLMs balance information from each modality, revealing varying degrees of unimodal collapse.
\citet{parcalabescu2025do} examine how VLM decoders rely on visual and textual inputs differently when producing answers versus explanations, noting greater reliance on text overall but an increased role for images in explanation tasks. Unlike these approaches, which focus on measuring how images and text are combined in general, our work studies a specific two-step reasoning process—entity identification followed by factual retrieval—and highlights a distinct performance gap when relying on visual instead of textual cues.
\citet{zhu2024unravelingcrossmodalityknowledgeconflicts} propose a pipeline to detect and mitigate what they refer to as parametric knowledge conflicts in large vision-language models, introducing a dynamic contrastive decoding method that improves performance on InfoSeek. While their work addresses cross-modal inconsistencies by focusing on parametric conflicts, we contribute a distinct perspective centered on bridging entity identification and factual retrieval.

Interestingly, when compared to human behavior, the behavior of VLMs we find in this paper contrasts with findings from cognitive psychology. Specifically, humans tend to perform better with visual rather than textual information, a phenomenon known as the picture superiority effect \citep{jenkins1967differential}. While explanations for this effect are still debated, the dual-coding theory \citep{PAIVIO} attributes this to images being encoded both visually and verbally, unlike text which is only encoded verbally. These findings suggest that humans benefit more from visual input, in contrast to VLMs, which are likely biased toward text due to its prevalence in their training data. We leave a further exploration of the relation between the two to future research.

\section{Conclusion}

This work explores how VLMs process visual entities, enhancing understanding of multi-modal architectures. Our findings highlight that reasoning about entities shown in an image is not a straightforward extension of reasoning about entities described in text, rather, it more closely resembles the processing of multi-hop questions. Across different model and visual backbones we see that models dedicate a substantial portion of their computational "budget" to entity identification and information processing, leaving few layers for the subsequent factual reasoning step. Thus, critical image-related processing often concludes 
relatively late in the processing flow, limiting the model's capacity to extract factual knowledge in the remaining layers thus limiting its overall performance. 
We also find that this issue persists even for entities that could be identified earlier, indicating that current models may not be using their representational capacity efficiently.

By highlighting these challenges, our results offer pathways for improvement. Potential directions for tackling the root of this issue include adjusting training regimes to better separate or balance identification and knowledge extraction steps, 
as well as designing architectural enhancements that reduce entanglement between these stages, which could lead to VLMs that better, and more reliably, utilize their internal knowledge. For instance, this behavior may be mitigated to some extent by using early-fusion models, that aim to create a shared embedding space between the visual and textual modalities. We leave this exploration for future work.

\section*{Limitations}
The scope of this work focuses on understanding the cause of the observed behavior, and leaves the solutions as future works.

In the second part of our experiments, we focused on the performance of \lv{} style models, as it is a canonical architecture in the field, and allows applying interventions with ease. While we did observe a gap in other architectures, we leave the probing experimentation on these for future work. Indeed, we did not test all type of VLMs, where some may exhibit different behavior to the kind described in this work. We believe that studying the gap we report in this work for a larger variety of VLM types would be an interesting direction for a future research.

When using visual representations, the image may contain features that can steer the model towards or away from the correct answer. For example, a football fan wearing his favorite team's shirt could be mistaken for a football player. This work does not go into disentangling the effects of such confounders on the observed gap.

Lastly, extending our analysis to entities with different properties and of additional ethnicities is an important direction for future work.

\section*{Acknowledgments}
We thank Omri Kaduri and Moran Yanuka for their input and feedback.
This work was supported by the Tel Aviv University Center for AI and Data Science (TAD) and the Israeli Science Foundation.

\bibliography{custom, anthology}

\appendix
\label{sec:appendix}

\section{Motivating Experiment - Results Breakdown By Type}
\label{sec:breakdown}

Tables \ref{tab:popvqa-extra} and \ref{tab:pval} show a breakdown of the results in section \ref{sec:gap} according to type. The first shows the image and text accuracy and drop, while the second shows a fine grained p-value analysis for the Wilcoxon signed-rank test. This breakdown provides some interesting insights. The drop in the paintings case is often low, with \qwen{} even doing better with the visual representation. This seems to be due to the model being able to provide more informed guesses for some of the questions. Other than that, there is no correlation between specific types and accuracy drop. It seems that different models have different strengths and weaknesses, for example \lms{} recognizes significantly less brand logos despite the name of the brand often appearing on the logo.

\begin{table}[ht]
\footnotesize
\centering

\begin{tabularx}{\columnwidth}{l|lcc|c}
\hline
 Type      & \#Identified & Img Acc & Txt Acc & \textbf{Drop} \\
\hline
 \multicolumn{5}{l}{\textbf{\ls{}}} \\
 \hline
    Brands    &  343 &     0.270 &     0.432 &  \textbf{0.162} \\
    Celebs    &  294 &     0.342 &     0.543 &  \textbf{0.201} \\
    Landmarks &  168 &     0.280 &     0.502 &  \textbf{0.222} \\
    Paintings &  120 &     0.128 &     0.299 &  \textbf{0.171} \\
\hline
 \multicolumn{5}{l}{\textbf{\lmc{}}} \\
 \hline
    Brands    &  364 &     0.238 &     0.374 &  \textbf{0.136} \\
    Celebs    &  128 &     0.278 &     0.379 &  \textbf{0.101} \\
    Landmarks &  212 &     0.250 &     0.486 &  \textbf{0.236} \\
    Paintings &  140 &     0.083 &     0.249 &  \textbf{0.165} \\
\hline
 \multicolumn{5}{l}{\textbf{\lms{}}} \\
 \hline
    Brands    &  237 &     0.169 &     0.406 &  \textbf{0.237} \\
    Celebs    &  136 &     0.301 &     0.454 &  \textbf{0.153} \\
    Landmarks &  149 &     0.259 &     0.459 &  \textbf{0.200} \\
    Paintings &  138 &     0.111 &     0.207 &  \textbf{0.096} \\
\hline
 \multicolumn{5}{l}{\textbf{\lb{}}} \\
 \hline
    Brands    &  416 &     0.546 &     0.630 &  \textbf{0.083} \\
    Celebs    &  415 &     0.564 &     0.742 &  \textbf{0.179} \\
    Landmarks &  346 &     0.534 &     0.655 &  \textbf{0.122} \\
    Paintings &  109 &     0.435 &     0.455 &  \textbf{0.020} \\
\hline
 \multicolumn{5}{l}{\textbf{\qwen{}}} \\
 \hline
    Brands    &   462  &     0.363 &     0.411 &  \textbf{0.049} \\
    Celebs    &   1240 &     0.418 &     0.544 &  \textbf{0.125} \\
    Landmarks &   1059 &     0.474 &     0.489 &  \textbf{0.015} \\
    Paintings &   382  &     0.445 &     0.276 & \textbf{-0.168} \\
\hline
\end{tabularx}
\caption{Fine-grained results on \pop{} broken down by different entity types. Reported accuracy is first calculated per subject then averaged across the entire dataset.}
\label{tab:popvqa-extra}
\end{table}

\begin{table}[t!]
\footnotesize
\centering
\begin{tabularx}{\columnwidth}{l|ccc}
\toprule
Split & \ls & \lmc & \lms \\
\midrule
Overall & $3.83 \times 10^{-76}$ & $1.42 \times 10^{-49}$ & $6.36 \times 10^{-47}$ \\
Brands & $1.07 \times 10^{-21}$ & $1.16 \times 10^{-16}$ & $4.78 \times 10^{-21}$ \\
Celebs & $3.48 \times 10^{-40}$ & $1.34 \times 10^{-8}$  & $6.87 \times 10^{-12}$ \\
Landmarks & $4.88 \times 10^{-18}$ & $6.22 \times 10^{-22}$ & $2.90 \times 10^{-13}$ \\
Paintings & $2.51 \times 10^{-7}$  & $1.29 \times 10^{-8}$  & $7.33 \times 10^{-5}$ \\
\bottomrule
\end{tabularx}
\vspace{-0.1in}
\begin{tabularx}{\columnwidth}{l|ccc}
\toprule
Split & LLaVA-34B & Qwen2-VL \\
\midrule
Overall & $4.27 \times 10^{-56}$ & $4.02 \times 10^{-29}$ \\
Brands & $9.92 \times 10^{-11}$ & $7.60 \times 10^{-5}$ \\
Celebs & $3.42 \times 10^{-50}$ & $1.33 \times 10^{-115}$ \\
Landmarks & $3.59 \times 10^{-13}$ & $1.17 \times 10^{-3}$ \\
Paintings & $3.11 \times 10^{-1}$  & $1.00$ \\
\bottomrule
\end{tabularx}
\caption{Fine-grained p-value analysis of Wilcoxon signed-rank test between the per-entity accuracies for visual versus textual representations, broken down by different entity types.}
\vspace{-0.1in}
\label{tab:pval}
\end{table}

\section{Additional Experiment: Critical Information is Propagated Late}
\label{sec:knockout}
To investigate the stages of the model's processing pipeline where entity identification occurs, we ask the model to identify the subject in an image and employ an attention knockout strategy following the methodology of \citet{geva-etal-2023-dissecting}. This involves selectively prohibiting attention from textual tokens to image tokens at various layers of the model, effectively blocking cross-modal information flow. We do this by inserting $-inf$ in the attention mask at $[x, y]$ where $x$ are the indices of $\mathbf{h}_{1, t}^{\ell} \dots \mathbf{h}_{m, t}^{\ell}, \mathbf{h}_{1, g}^{\ell} \dots \mathbf{h}_{k, g}^{\ell}$, $y$ are the indices of $\mathbf{h}_{1, v}^{\ell} \ldots \mathbf{h}_{n, v}^{\ell}$, and $\ell$ is any layer we want to knock out. We conduct this experiment in two directions:

\noindent 1. \textbf{Top-down:} Knocking out attention from a start layer until the final layer of the model.

\noindent 2. \textbf{Bottom-up:} Knocking out attention from the model's first layer until an end layer.

Notably, part of this experiment has been done in concurrent work by \citep{kaduri2024whats}, but on different data, and asking different questions.
This experiment checks whether the VLM trend follows \citet{geva-etal-2023-dissecting} which analyze the information flow for factual recall given textual inputs.

\begin{figure}[t]
    \centering
    \includegraphics[width=1\linewidth]{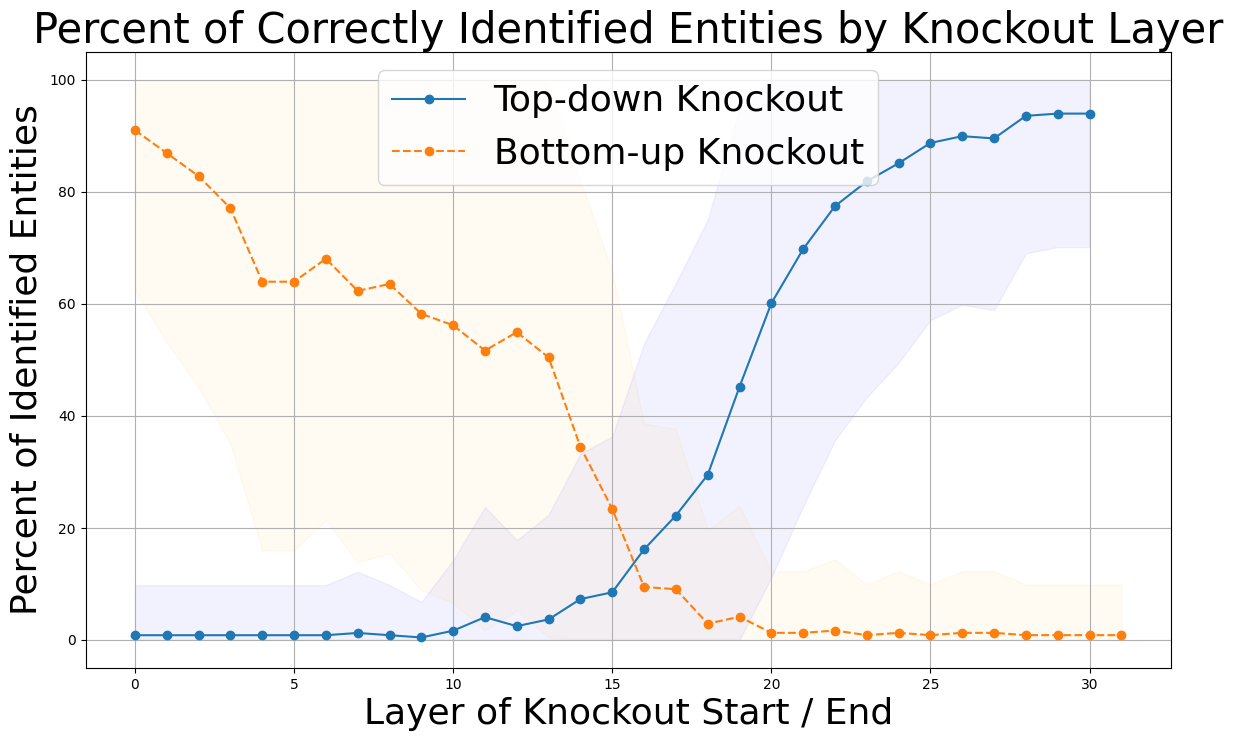}
    \caption{\textbf{Attention Knockout:} query and generated tokens are restricted from attending to the image tokens in all layers from layer 0 until the marker (in orange) or from the marker until the last layer (in blue)}
    \label{fig:knockout}
\end{figure}

\noindent \textbf{Results.}
Figure \ref{fig:knockout} presents both types of knockout on \ls{}:

\noindent 1. \textbf{Top-down:} Knocking out the second half or more of the model's attention results in complete collapse of model prediction. The sharp change in layers 15-20 indicates that these layers are critical for information flow from the image tokens to the query and generation tokens. Knocking out the last third of layers has a minimal effect on the model's output, indicating the model no longer attends to the visual tokens after the critical mid-layer processing is complete, demonstrating a transition to text-dominated reasoning.

\noindent 2. \textbf{Bottom-up:} Shows that unlike the top layers, restricting the attention in early layers has an immediate and progressive effect on the prediction, underscoring their importance. The effect increases around the middle layers until again collapsing the model's prediction. This shows that there could be some redundancy in the first layers, and the model can begin attending to image tokens quite late in the processing pipeline and still achieve mostly correct entity identification.

These trends mostly agree with \citet{geva-etal-2023-dissecting} that knock out attention edges of subject tokens in LLMs. This may indicate that the model treats the image tokens similarly to subject textual tokens. One must remember that unlike the textual tokens, the image tokens are already processed and enriched within the vision encoder. This means that although one could hope that image tokens would enjoy extraction at much earlier stages, allowing for contextualization by the text tokens through more layers, the model exhibits suboptimal integration of image information, relying heavily on mid-layer processing while under utilizing earlier layers for cross-modal fusion.

\noindent \textbf{Qualitative results.}
To complement the quantitative results, we examined qualitative examples of this degradation, in bottom-up knockout. For instance, Table \ref{tab:examples} shows the models output for different values of $\ell$ on an image of LeBron James in action. This example aligns with the general observed quantitative trend, showing that entity identification degrades progressively when employing bottom up attention knockout, the model continues to extract useful visual features before ultimately failing and hallucinating. Another such example at the end.

\begin{table}[t!]
\footnotesize
\centering
\begin{tabularx}{\columnwidth}{lX}
\toprule
\hspace{-0.1in} Layer & Output \\
\midrule
10 & LeBron James is an American professional basketball player who was born on December 30, 1979, in Akron, Ohio \\
19 & The image features a man wearing a yellow shirt and a red hat, holding a basketball in his hands. He appears to be a basketball player \\
20 & The image features a man wearing a yellow shirt, standing in front of a crowd of people. \\
21 & The image features a man wearing a suit and tie, standing in front of a building. \\
22 & The image features a man named LeBron James, who is an American professional basketball player. \\
29 & The image features a man named "Mike" who is standing in a field. He is wearing a black shirt \\
\bottomrule
\end{tabularx}
\caption{An example of bottom-up knockout on an image of LeBron James. The layer numbers indicate the final knockout layer. Only layers with meaningful change in output were included for brevity. Note that the model briefly recovers accurate entity identification.}
\label{tab:examples}
\end{table}

\section{Additional Qualitative Examples of Attention Knockout}
\label{sec:supp}

Bellow is another result of bottom up attention knockout. Every line shows the output of the model after knocking out another layer.
\begin{enumerate}
    \item \texttt{Donald Trump is the 45th President of the United States, serving from January 2017 to January 2021.}
    \item \texttt{Donald Trump is an American businessman, politician, and television personality. He was born on June 14, 1946}
    \item \texttt{Donald Trump is an American businessman, politician, and television personality. He was born on June 14, 1946}
    \item \texttt{Donald Trump is an American businessman, politician, and television personality. He was born on June 14, 1946}
    \item \texttt{The image features a man wearing a suit and tie, standing in front of a microphone. He appears to be a public speaker or a politician}
    \item \texttt{The image features a man wearing a suit and tie, standing in front of a microphone. He appears to be giving a speech or a presentation}
    \item \texttt{The image features a man wearing a suit and tie, standing in front of a microphone. He appears to be giving a speech or a presentation}
    \item \texttt{The image features a man wearing a suit and tie, standing in front of a microphone. He appears to be giving a speech or a presentation}
    \item \texttt{The image features a man with a receding hairline, wearing a suit and tie, and giving a speech.}
    \item \texttt{The image features a man with a distinctive hairstyle, sitting at a podium and giving a speech.}
    \item \texttt{The image features a man with a distinctive hairstyle, sitting at a podium and giving a speech.}
    \item \texttt{The image features a man with short hair, wearing a suit and tie, and giving a speech. He appears to be a professional orator}
    \item \texttt{The image features a man with a distinctive hairstyle, wearing a suit and tie.}
    \item \texttt{The image features a man with short hair, wearing a suit and tie, and giving a speech. He appears to be a professional orator}
    \item \texttt{The image features a man with a beard, wearing a suit and tie, and giving a speech.}
    \item \texttt{The image features a man wearing a suit and tie, standing in front of a microphone. He appears to be a public speaker}
    \item \texttt{The image features a man wearing a suit and tie, standing in front of a microphone. He appears to be a public speaker}
    \item \texttt{The image features a man wearing a suit and tie, standing in front of a blue background. He appears to be a professional or businessman}
    \item \texttt{The image features a man wearing a suit and tie, standing in front of a microphone. He appears to be a professional speaker or an announcer}
    \item \texttt{The image features a man wearing a suit and tie, standing in front of a microphone. He appears to be a professional speaker or an announcer}
    \item \texttt{The image features a man wearing a suit and tie, standing in front of a building. He appears to be a businessman or a professional}
    \item \texttt{The image features a man wearing a suit and tie, standing in front of a building. He appears to be a businessman or a professiona}
    \item \texttt{The image features a man standing in front of a building, possibly a church. He is wearing a suit and tie}
    \item \texttt{The image features a man standing in front of a building, possibly a church. He is wearing a suit and tie}
    \item \texttt{The image features a man standing in a field, possibly a farmer or a worker. He is wearing a hat}
    \item \texttt{The image features a man standing in a field, possibly a farmer or a worker in the field. He is wearing a hat}
    \item \texttt{The image features a man standing in a field, possibly a farmer or a worker in the field. He is wearing a hat}
    \item \texttt{The image features a man standing in a field, possibly a farmer or a worker in the field. He is wearing a hat}
    \item \texttt{The image features a man standing in a field, possibly a farmer or a worker in the field. He is wearing a hat}
    \item \texttt{The image features a man named "Mike" who is wearing a suit and tie. He appears to be a businessman}
    \item \texttt{The image features a man named "Mike" who is wearing a suit and tie. He appears to be a businessman}
    \item \texttt{The image features a man named "Mike" who is wearing a suit and tie. He appears to be a businessman}
\end{enumerate}

\section{Resources}
\label{sec:resources}

All the experiments were conducted using a single A100 80GB or H100 80GB GPU. 

\section{Ethics of Data Collection}
\label{sec:ethics}

Note that the data was collected with only permissive licenses. Moreover, we have an approval from our institution IRB for "Checking and understanding the performance of deep neural networks".

\end{document}